\documentclass{article}

\usepackage{arxiv}

\usepackage[utf8]{inputenc} % allow utf-8 input
\usepackage[T1]{fontenc}    % use 8-bit T1 fonts
\usepackage{hyperref}       % hyperlinks
\usepackage{url}            % simple URL typesetting
\usepackage{booktabs}       % professional-quality tables
\usepackage{amsfonts}       % blackboard math symbols
\usepackage{nicefrac}       % compact symbols for 1/2, etc.
\usepackage{microtype}      % microtypography
\usepackage{cleveref}       % smart cross-referencing
\usepackage{lipsum}         % Can be removed after putting your text content
\usepackage{graphicx}
\usepackage{natbib}
\usepackage{doi}

\usepackage{cite}
\usepackage{textcomp}
\usepackage{xcolor}
\usepackage{subcaption}
\usepackage{multicol}
\usepackage{multirow}
\usepackage{float}
\usepackage{adjustbox}
\usepackage{booktabs}
\usepackage{authblk}
\newcommand{\review}[1] {{\color{black} #1}}
\newcommand{\changed}[1] {{\color{black} #1}}
\newcommand{\blue}[1]{$_{\downarrow #1}$}
\newcommand{\red}[1]{$_{\uparrow #1}$}

\title{One Period to Rule Them All: Identifying Critical Learning Periods in Deep Networks}

\date{}
\author[1]{Vinicius Yuiti Fukase$^{*}$}
\author[1]{Heitor Gama$^{*}$}
\author[1]{Barbara Bueno}

\author[1]{\\Lucas Libanio} % Line break before Lucas
\author[1]{Anna Helena Reali Costa}
\author[1]{Artur Jordao}

\affil[1]{Escola Politécnica, Universidade de São Paulo, Brazil}

\hypersetup{
pdftitle={One Period to Rule Them All: Identifying Critical Learning Periods in Deep Networks},
pdfsubject={},
pdfauthor={},
pdfkeywords={Deep Learning, Neural Networks, Critical Periods, Regularization, Green AI.},
}

\begin{document}
% Switch to symbol footnotes for title
\renewcommand{\thefootnote}{\fnsymbol{footnote}}
\maketitle
\footnotetext[1]{Equal contribution.}
\renewcommand{\thefootnote}{\arabic{footnote}}  % Revert back to normal numbering
\setcounter{footnote}{0} % Reset footnote counter for main text

% \begin{abstract}
% 	\lipsum[1]
% \end{abstract}

% % keywords can be removed
% \keywords{First keyword \and Second keyword \and More}
\begin{abstract}
Critical Learning Periods comprehend an important phenomenon involving deep learning, where early epochs play a decisive role in the success of many training recipes, such as data augmentation. Existing works confirm the existence of this phenomenon and provide useful insights. However, the literature lacks efforts to precisely identify when critical periods occur.
%
% regularization |  into it
In this work, we fill this gap by introducing a systematic approach for identifying critical periods during the training of deep neural networks, focusing on eliminating computationally intensive regularization techniques and effectively applying mechanisms for reducing computational costs, such as data pruning. Our method leverages generalization prediction mechanisms to pinpoint critical phases where training recipes yield maximum benefits to the predictive ability of models. By halting resource-intensive recipes beyond these periods, we significantly accelerate the learning phase and achieve reductions in training time, energy consumption, and CO$_2$ emissions. Experiments on standard architectures and benchmarks confirm the effectiveness of our method. Specifically, we achieve significant milestones by reducing the training time of popular architectures by up to 59.67\%, leading to a 59.47\% decrease in CO$_2$ emissions and a 60\% reduction in financial costs, without compromising performance.
Our work enhances understanding of training dynamics and paves the way for more sustainable and efficient deep learning practices, particularly in resource-constrained environments. In the era of the race for foundation models, we believe our method emerges as a valuable framework. The repository is available at~\url{https://github.com/baunilhamarga/critical-periods}.
%%Our methodology leverages novel metrics, including Layer Rotation and Gradient Confusion, to pinpoint critical phases where regularization yields maximum generalization benefits. 
\end{abstract} 
\keywords{Deep Learning \and Neural Networks \and Critical Periods \and Regularization \and Green AI.}

\section{Introduction} 
Critical learning periods refer to a deep learning phenomenon where early epochs determine success in many training recipes, including regularization and the capacity of the model to combine information from diverse sources~\citep{Achille:2019,kleinman2023critical}. Understanding critical learning periods in deep learning can significantly boost training efficiency and overall model performance. Despite their acknowledged significance, accurately identifying \changed{such} periods during the training phase remains elusive. However, these periods significantly influence the dynamics and effectiveness of learning. Thus, systematically identifying them is essential to optimize training efficiency and model generalization.

Prior research confirms that critical periods manifest early in the training process, beyond which numerous training methods yield minimal to no additional advantage~\citep{Achille:2019}. \citet{kleinman2023critical} emphasize the capacity of neural networks to combine data from varied origins, significantly hinging on their exposure to appropriately correlated stimuli during initial training stages. Intricate and unpredictable early transient dynamics illustrate the emergence of critical periods. These dynamics play a crucial role in determining final efficacy and the nature of representations acquired by the system upon completion of training. The seminal work by~\citet{kleinman2024critical} demonstrates that learning ability does not increase monotonically during training and that a memorization phase occurs in early epochs, where models retain most discriminative information. Furthermore, \changed{the authors} show that critical periods occur \changed{during the} memorization phase and, after this, the model may start to weaken its ability to retain knowledge (forgetting phase). 

%\review{This suggests that the value of regularization does not come from directing the model towards critical periods with superior generalization post-critical period.}

Existing studies argue that the effectiveness \changed{of} regularization during model training extends beyond merely preventing solution entrapment in local minima~\citep{Achille:2019}. Specifically, they find that adjusting regularization practices after a critical period in training changes weight values and thereby the position of the model in the loss landscape. Consistent generalization confirms that this adjustment does not improve the predictive ability of models. Rather, its importance lies in guiding initial stages of training towards areas of the loss landscape that are rich in diversity, yet equally effective solutions with strong generalization characteristics. Although these works \changed{play} an important role in critical learning periods, none of them offer a systematic way to identify critical periods. It is worth mentioning that~\citet{golatkar2019time} find that critical periods emerge early in training but not exactly when (i.e., the epoch). Therefore, a natural question that arises is:

\begin{quote}
\centering  
\emph{How to identify critical periods during the course of training?}
\end{quote}

Given calls for more efficient training in the era of foundation models~\citep{Liu:2024}, answering this question yields two notable gains. Firstly, we could apply strong regularization only while the model is apt to absorb information (before critical periods). Secondly, we could reduce \changed{the number} of training examples through data pruning only after the critical period; thus, preventing loss in predictive ability. Together, these strategies enable better allocation of computational resources, avoiding unnecessary demands, and significantly speeding up the training process.

\noindent
\textbf{Research Statement and Contributions}. To sum up, our work has the following research statement. \emph{Throughout the course of training, a simple generalization estimation enables successfully identifying when the critical period emerges. Adjusting training recipes at this point --  such as stopping data augmentation or refining data pruning -- preserves predictive ability, while significantly reducing training costs.} 
Among our contributions, we highlight the following. 1) We emphasize the importance of discovering the moment (epoch) at which the critical periods emerge. 
2) We introduce a systematic approach for identifying such a moment during the training process. 3) Across various benchmarks and architectures, we show that our method significantly reduces computational costs by eliminating training recipes after critical periods. This efficiency comes with a minimal trade-off in accuracy. Overall, our contributions not only enhance understanding of the training dynamics, but also offer a practical tool for optimizing resource use in deep learning. 

Extensive experiments across a broad range of benchmarks (CIFAR-10/100~\citep{cifar10,cifar100}, EuroSat~\citep{helber2018eurosat}, Tiny ImageNet~\citep{Le2015tinyimagenet} and ImageNet30~\citep{hendrycks2019imagenet30}) and architectures confirm our research statement and contributions. Specifically, 
we reduce training time by up to 59.67\% with a negligible accuracy drop. In terms of Green AI~\citep{lacoste2019quantifying,faiz2023llmcarbon,morrison2025holistically}, these results represent a significant advancement in minimizing carbon emissions associated with energy use during model deployment. In particular, we reduce CO$_2$ emissions by 59.47\% and financial costs by 60\%.
%\todo{Coloque algum valor quantitativo que chame atenção como fizeram acima}

% We investigate various metrics to identify critical learning periods, and discover that Layer Rotation stands out as the most effective. Therefore, we now focus our efforts on Layer Rotation, which leads us to gain deeper insights and find strong evidence of its efficiency in pinpointing critical learning periods.

\section{Related Work}

\noindent
\textbf{Critical Periods.} The training dynamics of neural networks reveal that early stages of learning play a decisive role in determining model performance~\citep{Achille:2019,golatkar2019time,maini2023neural}. These stages, named critical periods, \changed{represent} phases where regularization techniques, such as weight decay and data augmentation, have the most significant impact on generalization~\citep{golatkar2019time,Achille:2019}. Once these periods pass, persistently applying regularization \changed{yields} diminishing returns, adding computational overhead without comparable benefits~\citep{kleinman2023critical}.

Recent studies emphasize the importance of understanding and leveraging \changed{critical learning periods.}  
%Metrics like gradient confusion and layer rotation have been proposed to capture dynamic shifts during training, providing potential indicators of these phases~\citep{sankararaman2020impact,carbonnelle2019layer}. 
For example,~\citet{golatkar2019time} and~\citet{Achille:2019} suggest that reducing or even removing regularization after initial learning phases can lead to more effective training. 
%Furthermore, while we apply our approach in standard deep learning scenarios, recent works explore the role of critical learning periods beyond traditional deep machine learning, particularly in federated learning. For instance, Yan et al.~\citep{yan2022seizing,yan2023defl,yan2023criticalfl} investigate how leveraging critical periods can enhance robustness against adversarial attacks and optimize client selection strategies in federated settings.
Recent works explore the role of critical learning periods beyond traditional deep machine learning, particularly in federated learning. For instance,~\citet{yan2022seizing,yan2023defl,yan2023criticalfl} investigate how leveraging critical periods can enhance robustness against adversarial attacks and optimize client selection strategies in federated settings.
From the lens of combining sources of information,~\citet{kleinman2023critical} observed that critical periods also impair the ability to synergistically merge data across multiple sources.
%observed that critical periods damage the model ability to harvest synergistic information among multiple sources of data.
Although these studies provide insights into critical periods phenomena, mainly from the theoretical perspective, none suggest a systematic method for identifying them.
Importantly, identifying and acting upon critical periods offers a promising direction for reducing computational cost and improving training efficiency~\citep{faiz2024llmcarbon}. 
Our work fills this gap and introduces an effective strategy for pinpointing critical periods. In practical terms, this enables halting training recipes at optimal points, achieving comparable or improved accuracy while significantly reducing training time.

\noindent
\textbf{Generalization Estimation.} Estimating how neural networks generalize to unseen data early in the training process is crucial for both practical applications and for advancing the theoretical understanding of deep models~\citep{Ballester:2024}.
For example,~\citet{sankararaman2020impact} estimate training convergence using the gradient confusion among batches of samples during the SGD updates. Similarly,~\citet{Chen:2023} estimate training dynamics according to the stability of gradient directions across batches and parameter updates.
From a different perspective,~\citet{carbonnelle2019layer} introduce a simple yet effective indicator of generalization. Their method, named \emph{Layer Rotation}, estimates generalization performance at a given training epoch taking into account the cosine distance between its weights and those from the random initialization.

As we shall see, our method leverages the metrics above to discover critical periods. Throughout our analysis, however, we observe that the methods by~\citet{sankararaman2020impact} and~\citet{Chen:2023} become unreliable for this purpose as they reveal inconsistent relationships with model accuracy, exhibit significant noise or have high computational cost.
%Gradient predictiveness~\citep{Chen:2023},, which measures the stability of gradient directions across batches and parameter updates, has been explored as an indicator of training dynamics. We tested a variation of this approach by measuring the cosine similarity of gradients calculated on fixed batches across consecutive epochs. However, the results revealed no consistent relationship with model accuracy and showed significant noise, making it unreliable for identifying critical periods. 
%%
%\changed{Overall, metrics like gradient confusion and layer rotation try to capture dynamic shifts during training and provide potential indicators of the generalization behavior~\citep{sankararaman2020impact,carbonnelle2019layer}.}

% Heitor - Eu queria deixar claro nessa seção abaixo que nosso "augmentation" será referenciado como aumento de quantidade de amostras e não a simples transformação. Isso é para não haver confusão quando falamos de "stop augmentation": o leitor deve entender que ainda haverá transformações aleatórias, mas não haverá aumento de quantidade de amostras.
% Heitor - update: eu já deixei claro na seção 'Experimental Setup', então talvez pode deixar só lá.
\noindent
\textbf{Data Augmentation.} The current paradigm for solving cognitive tasks using deep learning involves training models on large amounts of data (i.e., web-scale data)~\citep{Dubey:2024}. Modern models, such as Llama 3, reinforce that the secret ingredient behind positive results lies in web-scale and high-quality data~\citep{Dubey:2024}. In this direction, data augmentation becomes one of the most important training recipes. It turns out that, due to the stochastic nature of state-of-the-art techniques, many of them allow the creation of multiple samples from a single one~\citep{yun2019cutmix}. Thus, it is possible to increase both data quantity and diversity without a labor-intensive and costly labeling process. For example, popular augmentations such as PixMix~\citep{Hendrycks:2022}, MixUp~\citep{Zhang:2018}, and Cutout~\citep{Zhong:2020} apply transformations to the original sample with a given probability $p$.
From this perspective, one could generate \changed{arbitrarily large training datasets} just using data augmentations $k$ times per input. 
%As a concrete example, in Appendix~\ref{sec:gradient_confusion_experiment}, we expand the original dataset by applying random image transformations three times per input and improve the final accuracy by $1$ percentage point compared to the original dataset. 

Regardless of whether the increase in data quantity stems from collection or data augmentation, handling more data \changed{incurs high computational, energy, and financial costs.}
% Combining these complementary techniques offers a pathway to maximize training efficiency while balancing resource constraints.

\noindent
\textbf{Data Pruning.} While data augmentation focuses on enriching the dataset, data pruning aims to reduce computational costs by selecting a smaller, representative subset of data, without sacrificing predictive performance.
%addresses these challenges by reducing dataset size without sacrificing performance.
%Data pruning reduces computational costs by selecting a smaller, representative subset of data while maintaining model performance. 
Existing methods typically fall into two categories. Importance-based approaches~\citep{Han:2023,Xia:2024,Choi:2024} evaluate the relevance of individual data points. Optimization-based techniques aim to retain the core characteristics of the original dataset~\citep{Mahabadi:2023,Engstrom:2024,xiao2025duoattention,li2025idiv}.
However, both approaches often involve complex and computationally expensive procedures. Interestingly,~\citet{Okanovic:2024} demonstrate that well-designed random pruning strategies can match or even outperform many sophisticated methods, underscoring the potential of simpler, more scalable solutions. 
% To mitigate accuracy loss during pruning, Qin et al.~\citep{Qin:2024} proposes updating gradients dynamically, achieving near-lossless performance without compromising training efficiency.

Despite positive advancements, most pruning techniques overlook the critical learning periods that occur early in training, when data selection may exert a disproportionate impact on model performance. To address this gap, we propose applying data pruning only after our method identifies the critical period. This approach ensures that sufficient data supports the early stages of learning, allowing even random pruning methods -- like the one suggested by~\citet{Okanovic:2024} -- to achieve superior performance with reduced complexity.

According to our results, this strategy reduces the overall training time by up to $2.5\times$ without degrading generalization. Additionally, unlike \changed{computationally} intensive techniques, our proposed form of data pruning incurs no additional costs.

\section{Preliminaries and Proposed Method}\label{sec::methodology}
\noindent
\textbf{Preliminaries.} Assume $X$ and $Y$ a set of training samples and their respective class labels (i.e., categories). Let $\mathcal{F}(\cdot, \theta)$ be a neural network parameterized by a set of weights $\theta$. From a random initialization $\theta^0$, an iterative process (e.g., SGD) updates $\theta^i$ towards a minimum of \changed{the} loss function $\mathcal{L}$, where $i$ indicates the $i$-th iteration of this process. 
%$X$ and $Y$ on the traditional supervised paradigm. 
Unless stated otherwise, we conduct the iterative update process of $\theta$ across $N$ training epochs.
%Our experiments conduct iterative processes across $N$ training epochs. %\todo{Isso precisa se encaixar na sentença anterior}
%\todo{Precisamos definir o $N$ que são as épocas de treinamento}
%Seja $\mathcal{F}(\cdot, \cdot)$ uma rede neural que será treinada usando o paradigma supervisionado sob os dados $X$ e os rótulos $Y$. 
%
%Defina $\theta_i$ os parâmetros de $\mathcal{F}(\cdot, \cdot)$ que serão ajustados por um processo iterativo de otimização (ex. \emph{Stochastic Gradient Descent} -- SGD), onde $i$ representa uma iteração (época) deste processo. 
%

To improve generalization of $\mathcal{F}$, previous works \changed{typically} apply regularization and data augmentation mechanisms during the optimization \changed{iterative} process~\citep{Cubuk:2020,Hendrycks:2022,kim2025controllable,robine2025simple}. Particularly, in this work, we focus on regularization through data augmentation techniques, as we formalize below.

Let $T(\cdot)$ be a function that receives samples from $X$ and modifies its content, producing a new set of same size (i.e., $|T(X)| = |X|$). Typically, modern data augmentation methods incorporate stochastic elements enabling the creation of arbitrarily large datasets by applying $T$ multiple times~\citep{Cubuk:2020,Hendrycks:2022,kim2025controllable}. Formally, we can augment the original dataset by applying $T$ $k$ times\changed{, denoted by} $\{(T(X), Y)\}^k$. %
To simplify the notation, let $\mathcal{D} = (X, Y)$ represent the pair of samples and their respective labels. Thus, after performing data augmentation $k$ times, it is possible to rewrite data augmentation as $\{\mathcal{D}\}^k$.

%\changed{From the lens of data augmentation and using the formalism above, we can formalize the parameter update $\theta$ as follows:}
By applying data augmentation techniques, we can formalize the iterative process of updating $\theta$ as follows:
\begin{equation}\label{eq::SGD_batch}
	\theta^{i+1} = \theta^{i} - \eta \frac{1}{B} \sum_{b=1}^{B}\nabla\mathcal{L}(\{\mathcal{D}\}^k_b, \theta^i),
\end{equation}

where $B$ indicates the \emph{batch size}, $\mathcal{D}^k_b$ is a \emph{batch} of $b$ samples from augmented data, $\nabla\mathcal{L}$\footnote{It is possible to rewrite the gradient as follows: $ \mathcal{L}(\{(Y_b, \mathcal{F}(T(X_b)))\}^k, \theta^i)$, where $X_b$ and $Y_b$ are data \emph{batches} and respective labels.} corresponds to the gradient of the loss function with respect to the parameters $\theta^i$ and $\eta$ denotes the update magnitude (learning rate).

% \todo{Colocar que a partir dessa eq. o data pruning pode ser expresso usando k< 1 e T(.) sendo a identidade.}
% 
%A partir do formalismo acima, o objeto de pesquisa deste projeto consiste em encontrar uma época de treinamento $i$ de tal forma que seja possível reduzir a quantidade de aumento de dados, isto é, diminuir o valor $k$ (idealmente utilizar $k=1$). 
Building on the previous formalism, \changed{the end of a} critical learning period is a training epoch $i^* \in \{0, \ldots , N\}$ from which point onward different training recipes provide little or no benefit to generalization. Therefore, our goal is to identify $i^*$ in a way that effectively \changed{minimizes}, or ideally \changed{eliminates}, computationally intensive training recipes. Moreover, we can apply \emph{data pruning} (i.e., \( k < 1 \)) to further reduce the computational demands of the training phase. In this context, we adopt the method proposed by~\citet{Okanovic:2024}. Their work emphasizes that repeated random sampling is a simple yet effective method that can outperform more complex techniques. Hence, this method becomes attractive \changed{as} it incurs no additional costs while introducing variability and enhancing generalization. At each epoch, we apply \review{dynamic random data pruning by randomly selecting a} percentage of the training dataset. Therefore, the training data changes with every epoch, ensuring the model sees different data points. Formally, at each epoch \( i \), we select a random subset \( D_i \subset D \) of the training data, where \( D \) is the full training set and \( D_i = k \cdot D \)\changed{, where \( k \) represents the proportion of the dataset selected}. It is important to note that our training phase employs this process only when we consider data pruning mechanisms.
%data pruning and minimize, or ideally eliminate, computationally intensive training recipes such as data augmentation. 

In summary, by discovering $i^*$, we notably \changed{speed up} the overall training process while preserving predictive ability.
% Incluir citação
It is worth emphasizing that previous studies confirm the possibility of successfully reducing or eliminating training recipes after an iteration $i$~\citep{Achille:2019,golatkar2019time}. To the best of our knowledge, however, there are no efforts to determine \emph{when} to reduce it.

% Heitor - O T(.) aqui redefine a transformação de data augmentation que já havíamos definido anteriormente. Precisa mudar aqui ou lá para não haver inconsistência.
% \review{In the context of data pruning, from Equation~\ref{eq::SGD_batch} and considering that \(k < 1\), we can refine the data pruning process in terms of \(k\) and \(T(\cdot)\), where \(T(\cdot)\) denotes the identity function. This refinement allows us to define the pruned dataset \(D'\) more precisely as follows:  
% \[D' = \{(x, y) \in D \,|\, k \leq T(C(x, y))\}.\]

% Given \(k < 1\), this condition emphasizes a strict selection criterion, including only those data samples \((x, y)\) in \(D'\) whose criterion function \(C(x, y)\) value meets or surpasses the threshold \(T(\cdot)\) sets. This method effectively filters the dataset, retaining elements that significantly boost model performance or relevance according to the constraints involving \(k\) and \(T(\cdot)\).}

\noindent
\textbf{Proposed Method.} To grasp the intriguing generalization properties present in deep neural networks, it is \changed{crucial} to identify numerical indicators of generalization performance that \changed{remain applicable} across diverse training settings. In this context,~\citet{carbonnelle2019layer} propose a groundbreaking approach to understanding and improving neural network generalization, named \textit{Layer Rotation}. This approach \changed{focuses on} tracking the evolution of \changed{the} cosine distance between each weight vector of each layer and its initial state throughout \changed{the} training process. Following~\citet{mason-williams2024what}, instead of computing the cosine similarity between each weight of a given layer (as originally suggested by~\citet{carbonnelle2019layer}), we concatenate all the weights composing the model $\mathcal{F}$ and linearize them to form a single vector. According to their work~\citep{mason-williams2024what}, this process enables a representation of the neural network parameters. For \changed{ease} of exposition, we \review{will} keep indicating the single vector representing all weights of $\mathcal{F}$ as $\theta^0$ (random initialization) and $\theta^i$ (with $i>0$).
%Specifically, Mason-Williams et al.~\citep{mason-williams2024what} shows that, by concatenation, it is possible to flatten weights to form a vector representation of the neural network parameters. Let \(\theta^{i}\) be the flattened weight tensor at optimization step \(i\) (with \(i_0\) corresponding to initialization); 

Given the previous definition, the cosine distance between \(\theta^0\) and \(\theta^{i}\) defines layer rotation at training epoch \(i\). 
%The formula below defines cosine distance as:
Thus, we estimate the layer rotation in terms of
\begin{equation}\label{eq::Layer_Rotation}
%\[  
Cosine Distance = 1 - \frac{\mathbf{\theta^{0}} \cdot \mathbf{\theta^{i}}}{\|\mathbf{\theta^{0}}\| \|\mathbf{\theta^{i}}\|}.  
%\]
\end{equation}

Through a comprehensive suite of experiments encompassing a broad spectrum of datasets, network architectures, and training regimes, we uncover a consistent pattern: \emph{larger layer rotations (i.e. as cosine distance between the final and initial weights increases) reliably predict enhanced generalization performance}.

In order to visualize layer rotation evolution during training, we track the cosine distance between the current weight vector of each layer and its initial state across various training steps. Upon analyzing these curves on a validation set, we notice a characteristic shape pattern emerging throughout the training process. To systematically identify critical periods within this evolution, we adopt an approach that performs linear regression over a window of 5 epochs ($w$). This choice proves itself effective in our experiments, emerging as the smallest window size that still allows capturing significant changes in \review{layer} behavior. A smaller window would make the analysis too vulnerable to minor fluctuations.

% We scale the number of epochs and cosine distance from 0 to the maximum value, 200 and 1, respectively
We scale the number of epochs and cosine distance from 0 to their respective maximum values. This way, we can graphically calculate the learning variation during training by the angle \(\alpha\) that indicates the rate of change within this window. Then, we determine its value by the linear regression of the normalized data and the arctangent calculation of the regression coefficient $m$ in degrees. To accomplish this, we define a set of data points $\{(u_1, v_1), (u_2, v_2), \ldots, (u_w, v_w)\}$, where $u_i$ represents the epoch (independent variable) and $v_i$ the cosine distance between $\theta^i$ and $\theta^0$ (dependent variable) -- the Layer Rotation\review{.} The slope $m$ of the regression line that best fits these data points (in the least squares sense) is: 

\begin{equation}\label{eq::learning_variation1}
%\[
  m = \frac{\sum_{i=1}^{w}(u_i - \bar{u})(v_i - \bar{v})}{\sum_{i=1}^{w}(u_i - \bar{u})^2}.
%\]
\end{equation}

\changed{We therefore present the formula for calculating the angle as follows:}

\begin{equation}\label{eq::learning_variation2}
%\[
\alpha = \frac{180}{\pi} \arctan(m).
%\]
\end{equation}

In our initial experiments, we observe that an angle \changed{of} \(45^\circ\) \changed{marks} a pivotal moment in the training of neural networks, where the shift from rapid learning to careful refinement and optimization occurs. Therefore, we use this value throughout our work.
%\changed{After establishing the benchmark angle at \(45^\circ\), we substantiate this value through validation, demonstrating its significance as a pivotal moment in the training of neural networks where the shift from rapid learning to careful refinement and optimization occurs.}
% Setting the benchmark angle at \(45^\circ\) lets us accurately identify the critical period, since that point marks a pivotal moment in training neural networks. It signifies \changed{moving from rapid learning to careful refinement and optimization,} making \(45^\circ\) a key indicator of training dynamics change. \todo{VINICIUS - achamos por validação}
% Setting the benchmark angle at \(45^\circ\), which matches the curve's knee point, lets us accurately identify the critical period, since that point marks a pivotal moment in training neural networks. 
% It signifies shifting from quick significant weight adjustments scenario to a slower one, \changed{moving} from rapid learning to careful refinement and optimization, making \(45^\circ\) a key indicator of training dynamics change.
% . This shift is vital as it moves
%
%\changed{
%To encapsulate our approach, Algorithm~\ref{alg:method} presents the complete pseudocode. This algorithm outlines the sequential steps for computing layer rotation and the corresponding learning variation, thereby ensuring a clear and reproducible implementation of our method.
%\input{sections/Algorithm.tex}
%}

\section{Experiments}
\noindent
\textbf{Experimental Setup.} In our experiments, we use training cycles of $200$ epochs and SGD optimizer~\citep{golatkar2019time,kleinman2024critical} with a learning rate that starts at $0.01$, and is divided by $10$ at epochs $100$ and $150$. Furthermore, every sample is transformed randomly before each epoch by a combination of horizontal flips, crops, rotations, and translations. We apply this basic transformation even after we reduce the augmentation factor $k$ to $1$ or less mid-training. Unless stated otherwise, our data augmentation consists of repeating each sample $k$ times, with $k=3$ (formalism given in Section~\ref{sec::methodology}).
By doing this repetition step before the random transformations, we ensure each copy is slightly different from the original. This is important because we want to simulate the effect of having more samples, not just simple copies.

Regarding the models and datasets, we use the popular Residual networks~\citep{he2016resnet} at different depths, and the CIFAR-10/100 benchmarks. Overall, we apply these experimental settings (model $\times$ datasets) for most experiments because they are common practices in the context of critical period and data pruning~\citep{Qin:2024,golatkar2019time,kleinman2023critical}.
However, to confirm the effectiveness of our method, we also evaluate it on large-scale datasets, including EuroSat~\citep{helber2018eurosat}, Tiny ImageNet~\citep{Le2015tinyimagenet}, and ImageNet30~\citep{hendrycks2019imagenet30}.

Throughout experiments and discussions, the term baseline refers to the model without removing training recipes. In other words, it means the model training on standard practices without any knowledge and intervention of critical periods.

\noindent
\textbf{Metrics and Evaluation.} To evaluate the effectiveness of our method in terms of improvements in computational cost and generalization, we introduce two key metrics:
\begin{itemize}
    \item Normalized Training Cost: This metric quantifies the computational demand of training relative to the baseline (training with initial $k=3$ -- see Equation~\ref{eq::SGD_batch} -- during all epochs). It normalizes the cost by combining the number of epochs and the number of data points used per SGD update. Overall, this metric accounts for dynamic changes in dataset size during the training process, particularly during critical periods.
    \item Accuracy Delta: This metric measures the variation in model accuracy compared to the baseline model. It enables assessing the trade-off between computational efficiency and performance when removing training recipes.
\end{itemize}

\noindent 
\textbf{Enhancing Generalization Through Repeated Augmentation.} 
%*****Alternatives para esse título: (2) The Role of Expanded Dataset by Repeating Augmentation; (3) The Effect of Repeated Augmentation on Enhancing Generalization
We start our analysis by illustrating the advantages of generating arbitrarily large datasets by applying the same data augmentation $k$ times per sample. As we mentioned before, this is possible due to the stochastic nature of modern augmentation strategies. Specifically, because they employ transformations (i.e., rotation or crop) with a given probability $p$. To this end, we increase the number of training samples by three times ($k=3$ in Equation~\ref{eq::SGD_batch}) and compare the accuracy when using the dataset with data augmentation without changing its size (i.e., $k=1$). For a fair comparison, we consider the same initialization.

On the ResNet32 architecture, we observe an improvement of roughly 1 percentage point (pp) while using the increased dataset ($k=3$) and a similar gain with a deeper, high-capacity architecture (ResNet86\footnote{Here, we avoid using the popular ResNet56 and ResNet110 to prevent any bias in our subsequent analysis.}).
From these findings, we confirm that an effective training recipe for improving generalization is simply to expand the dataset size by repeating the same data augmentation $k$ times. Additionally, we highlight the following key observations. First, although these improvements may seem small, modern and complex data augmentation methods achieve similar gains~\citep{Zhang:2018,Zhong:2020}. Second, despite improving generalization, the training time increases proportionally; for example, moving from $k=1$ to $k=3$ increases the training time by roughly three times. Most importantly, this setting becomes a potential candidate for exploring the practical benefits of our method in discovering critical periods. In particular, we can begin training a model on the expanded version of a dataset ($k=3$) and then reduce the dataset to its original size ($k=1$), or even use smaller versions ($k<1$ -- data pruning), after identifying the critical period. Therefore, adapting the training size through $k$ enables leveraging the best of both worlds: \emph{higher generalization and lower training time.}

\noindent 
\textbf{Revisiting Critical Periods.} In this experiment, we \changed{re-examine} the existence of critical periods. However, in contrast to previous works~\citep{kleinman2024critical,kleinman2023critical,Achille:2019}, we analyze it from the lens of \changed{potential values of} $i^*$, taking into account the compromise between accuracy and computational cost. Specifically, our main objective is to identify the end of the critical period, i.e., an early epoch $i^*$ where we reduce training recipes until the training is complete and final accuracy is sufficiently close to the baseline accuracy.
To achieve this, we create oracle models that reveal every possible accuracy outcome after reducing the augmentation factor $k$ from $3$ to $1$ at epoch $i$. It is worth mentioning that, after eliminating data augmentation at epoch $i$, the training continues until completing $200$ epochs. Therefore, each oracle model uses $k=3$ for the initial $i$ epochs and $k=1$ for the remaining $200-i$ epochs.
%after a complete training session with reduced data augmentation from $k=3$ to $k=1$ at epoch $i$, for each $i$ from $1$ to $200$.

\begin{figure}[!t]
    \centering
    \includegraphics[width=\columnwidth]{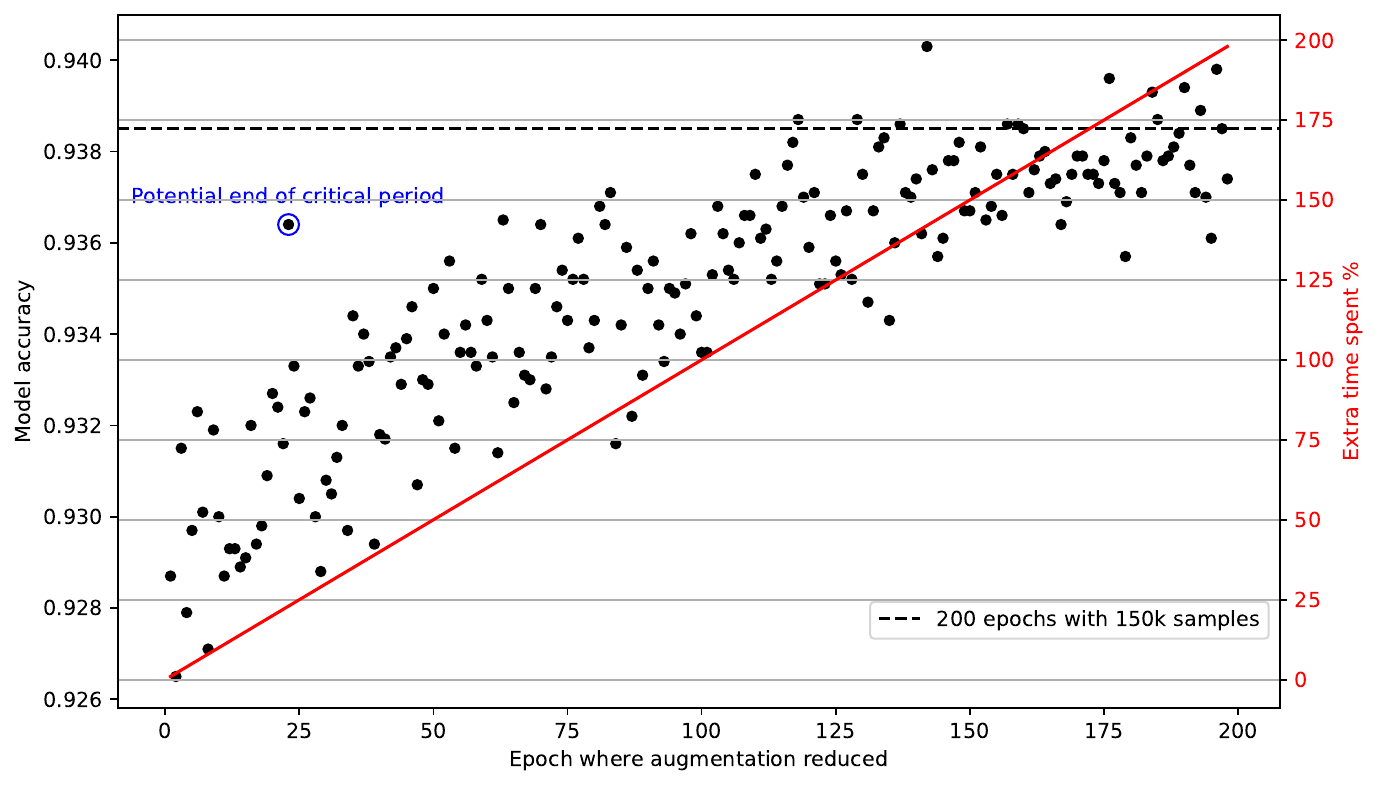}
    \caption{Critical period in ResNet32 training on CIFAR-10. Each point indicates a model with a complete training cycle of $200$ epochs. The figure shows the accuracy trends when data augmentation reduces from 150k to 50k samples per epoch at different epochs. The red line indicates the extra time cost.}
    \label{fig:resnet32_critical_periods}
	\vspace{-0.095cm}
\end{figure}

The aforementioned process produces $200$ oracle models and Figure~\ref{fig:resnet32_critical_periods} shows the accuracy of each one at the end of the training phase. Figure~\ref{fig:resnet32_critical_periods} reinforces the existence of critical periods early in training and supports that, after pinpointing $i^*$ (i.e. the end of the critical period), reducing augmentation is optimal in terms of accuracy and computational cost. Following this reasoning, epoch $24$ marks the potential end of the critical period. 
%It is important to note that only by epoch $60$ better models performance-wise emerge, but the accuracy gain is not significant enough to justify the extra computational cost.
It is important to note that better-performing models only emerge by epoch 60, but the accuracy gain is not significant enough to justify the additional computational cost.
%Appendix~\ref{sec:gradient_confusion_experiment} contains detailed experimental variables of these same settings on a \review{deeper architecture}.

\changed{Despite using checkpoints to avoid retraining initial epochs,} this experiment is very resource-intensive. For this reason, we choose CIFAR-10 and ResNet32, as both are computationally light while remaining sufficiently challenging and capable of delivering competitive performances for this task~\citep{he2016resnet, deng2020mcmc}.
%Figure~\ref{fig:resnet32_critical_periods} shows the results of this experiment.
%within our conditions. 
%começando com 150k amostras até a época do eixo x e a partir dai é 50k

From Figure~\ref{fig:resnet32_critical_periods}, the optimal epoch $i^*$ is explicit, however, this is the epoch we wish to identify without completing all these multiple models, allowing the decision to stop \changed{applying} training recipes at that point in a single run (i.e., on-the-fly during a single training phase). This is what our method aims to achieve.
Therefore, we now turn our attention towards automatically and systematically identifying $i^*$.

\noindent
\textbf{Effectiveness of Generalization Estimators.} Due to the nature of critical periods — epochs where training recipes promote significant impact on generalization~\citep{golatkar2019time,Achille:2019} — we argue that generalization estimation metrics can successfully identify $i^*$. Specifically, these metrics enable estimating the epoch where generalization begins to decline by analyzing their behavior among a range of epochs during training. \changed{In informal terms}, we believe these metrics can identify the circled point in Figure~\ref{fig:resnet32_critical_periods}.

Knowing the optimal $i^*$ beforehand from the previous experiment enables validating our argument to use generalization estimation metrics.
% For this purpose, \todo{we test metrics that indicate the proximity of $i^*$ on-the-fly to systematically detect the end of critical periods} without \changed{any \emph{prior} (and, unfortunately, unavailable)} knowledge.
%Knowing \changed{the optimal} $i^*$ beforehand from the previous experiment, we test metrics that indicate its proximity on-the-fly to systematically detect the end of critical periods without this \changed{\emph{prior} (and, unfortunately, unavailable)} knowledge.
%
Following previous studies, we explore the following metrics: Gradient Confusion~\citep{sankararaman2020impact}, Gradient Predictiveness~\citep{Chen:2023} and Layer Rotation~\citep{carbonnelle2019layer}. 
%We provide the formalism of Gradient Confusion and Gradient Predictiveness in Appendix~\ref{sec:gradient_confusion_experiment} and~\ref{sec:gradient_predictiveness_experiment}, respectively.

Regarding the first two metrics, we observe notable drawbacks. For example, Gradient Confusion is capable of accurately identifying critical periods (i.e., identify the epoch 24 in Figure~\ref{fig:resnet32_critical_periods}), but its computational cost is as intensive as the cost saved by reducing training recipes, making it impractical for real-time applications where reducing training time is a key objective. In contrast, Gradient Predictiveness identifies the critical period far from the optimal point. Particularly, this metric shows no correlation between its values and model accuracy on a validation set.
% From our experiments, we find that Layer Rotation is the most effective metric for this task overall.

Regarding Layer Rotation, we observe that as epochs progress, the cosine distance between the weights of each epoch and the initial weights increases. However, we notice that, at a certain point, the growth of this distance begins to diminish. \citet{carbonnelle2019layer} state that Layer Rotation achieves a network-independent optimum when the cosine distance of all layers reaches 1. However, our practical observations contradict this, revealing that distance values significantly fluctuate with the \review{architecture. Consequently, the pure application of this metric turns out to be architecture-dependent, making the Layer Rotation value arbitrary and unique to each architecture.}
% repeticao de architecture poderia ser evitada?
% \todo{Aqui é preciso colocar que em outras palavras usar o layer rotation puramente poderia ser ruim, e isso é o que nos motivou a plugar a equação 2 e 3 -- usar o histórico de épocas ao inves de uma época só. Mas ai tem que garantir um fluxo bom das ideias com a sentença seguinte}
Therefore, it becomes necessary to implement the learning variation during a window of epochs we introduce in Equations~\ref{eq::learning_variation1} and~\ref{eq::learning_variation2}. 
These changes allow analyzing the slope of the line generated by the linear regression of a 5-epoch window, evidencing the reduction in generalization over epochs (i.e., temporally). From these experiments, we identify epoch 24 as the turning point, at which the angle \(\alpha\) becomes less than 45º, indicating the transition of the curve to a smoother phase. 
The comparison with the oracle in Figure~\ref{fig:resnet32_critical_periods} highlights this point as having significant potential for the critical period.  Thereby, we confirm the potential of Layer Rotation in discovering the end of critical periods. 
%\changed{We emphasize that while some surrounding points (models) achieve similar generalization in later epochs, the computational cost incurred is not significant enough.} \review{Isso não é o mesmo que "Following this reasonin, epoch 24..." em azul?}

Besides accurately identifying the critical period, Layer Rotation is a quite efficient metric that demands negligible computational resources, as its calculations are simple and always based on comparing the current weights of the epoch with the initial ones (see Equation~\ref{eq::Layer_Rotation}). Therefore, we employ Layer Rotation as the main metric implemented in our systematic method.
%Although Layer Rotation exhibits better results, the other two metrics also provide key insights. On the other hand, we observe that 
% Thanks to these advantages, we confirm the effectiveness of Layer Rotation.
%Following the guidelines from Carbonelle et al.~\citep{carbonnelle2019layer}, we conduct experiments using the ResNet32 architecture and the CIFAR-10 dataset to validate their study. We observe that as epochs progress, the cosine distance between the weights of each epoch and the initial weights increases. However, we notice that at a certain point, the growth of this distance begins to diminish. Therefore, it becomes necessary to implement the changes described in Section 3. Implementing these changes allows us to analyze the slope of the line generated by the linear regression of a 5-epoch window, noting its reduction over the epochs. We identify epoch 24 as the turning point, at which the slope becomes less than 45º, indicating the transition of the curve to a smoother phase. The comparison with the oracle in Figure 1 highlights this point as having significant potential for the critical period.

\noindent
\textbf{Comparison with State-of-the-Art Data Pruning.} 
Existing data pruning methods provide positive results in reducing computational costs during the training of deep models~\citep{Qin:2024,Choi:2024}. However, to the best of our knowledge, no method currently takes into account the critical period phenomenon. 
	
In this experiment, we investigate the behavior of applying data pruning \emph{only after} our method identifies critical periods and compare the results with state-of-the-art methods. For this purpose, we build upon the experimental framework established by~\citet{Qin:2024}, using their evaluation of static and dynamic data pruning methods\footnote{Static data pruning removes a predetermined subset of data, with no reintroduction during training, whereas dynamic pruning allows for the subset to change as training progresses, adapting to the requirements of the model~\citep{Qin:2024}. Our method, as specified in Section~\ref{sec::methodology}, uses \changed{the dynamic} approach.} as a benchmark.  To this baseline, we integrate results from our approach, introducing data pruning  \changed{after identifying the critical periods during training.}. \changed{For a fair comparison}, we use ResNet18 as the backbone model, training it on CIFAR-10/100 datasets with pruning ratios of 30\%, 50\% and 70\%. It is worth mentioning our method diverges from traditional data pruning techniques by initially training the model on the full dataset (i.e $k=1$) for the first few epochs and then applying the pruning ratio from the critical periods onward. We compute the accuracy delta relative to a model trained for 200 epochs without data pruning, ensuring a fair comparison. Table~\ref{tab:pruning_comparison} shows the results.

\begin{table}[!t]
    \caption{Comparison of accuracy delta ($\Delta$Acc) \changed{with state-of-the-art} data pruning methods. \changed{Bold, underline, $\uparrow$ and $\downarrow$ mean the best results, second-best results, accuracy improvement and accuracy decrease, respectively.}}
    \centering
    \footnotesize
    \setlength{\tabcolsep}{3pt}
    \begin{adjustbox}{max width=0.45\textwidth, keepaspectratio}
    \begin{tabular}{cc|ccc|ccc}
    \toprule
    \multirow{3}{*}{} & Dataset  & \multicolumn{3}{c|}{CIFAR-10} & \multicolumn{3}{c}{CIFAR-100}  \\
    \midrule
    & Pruning Ratio \% & 30 & 50 & 70& 30 & 50 & 70 \\ \midrule
    \parbox[t]{4mm}{\multirow{17}{*}{\rotatebox[origin=c]{90}{Static}}}
    & Random
    & \blue{1.0} & \blue{2.3} & \blue{5.4} & \blue{4.4} & \blue{6.1} & \blue{8.5} \\
    & CD
    & \blue{0.6} & \blue{1.3} & \blue{4.8} & \blue{4.0} & \blue{5.9} & \blue{7.9} \\
    & Herding
    & \blue{3.4} & \blue{7.6} & \blue{15.5} & \blue{5.1} & \blue{6.4} & \blue{8.0} \\
    & K-Center
    & \blue{0.9} & \blue{1.7} & \blue{4.7} & \blue{4.1} & \blue{6.0} & \blue{8.0} \\
    & Least Confidence
    & \blue{0.6} & \blue{1.1} & \blue{5.3} & \blue{4.0} & \blue{5.9} & \blue{8.4} \\
    & Margin
    & \blue{0.7} & \blue{1.3} & \blue{4.7} & \blue{4.2} & \blue{6.0} & \blue{8.0} \\
    & Forgetting
    & \blue{0.9} & \blue{1.5} & \blue{3.9} & \blue{2.9} & \blue{5.1} & \blue{8.3} \\
    & GraNd-4
    & \blue{\underline{0.3}} & \blue{1.0} & \blue{4.4} & \blue{3.6} & \blue{6.8} & \blue{9.4} \\
    & DeepFool
    & \blue{0.5} & \blue{1.5} & \blue{5.6} & \blue{4.0} & \blue{5.0} & \blue{6.4} \\
    & Craig
    & \blue{0.8} & \blue{3.3} & \blue{7.2} & \blue{3.8} & \blue{6.3} & \blue{8.5} \\
    & Glister
    & \blue{0.4} & \blue{1.6} & \blue{4.7} & \blue{3.6} & \blue{5.0} & \blue{7.8} \\
    & Influence
    & \blue{2.5} & \blue{4.3} & \blue{7.3} & \blue{3.8} & \blue{6.2} & \blue{9.5} \\
    & EL2N-2
    & \blue{1.2} & \blue{2.4} & \blue{5.8} & \blue{4.1} & \blue{7.2} & \blue{9.7} \\
    & EL2N-20
    & \blue{\underline{0.3}} & \blue{0.5} & \blue{3.7} & \blue{1.0} & \blue{6.1} & - \\
    & DP
    & \blue{0.7} & \blue{1.8} & \blue{4.8} & \blue{1.0} & \blue{5.1} & - \\
    \midrule
    \parbox[t]{4mm}{\multirow{5}{*}{\rotatebox[origin=c]{90}{Dynamic}}}
    & Random
    & \blue{0.8} & \blue{1.1} & \blue{2.6} & \blue{0.9} & \blue{2.9} & - \\

    & $\epsilon$-greedy
    & \blue{0.4} & \blue{0.7} & \blue{\underline{1.5}} & \blue{1.8} & \blue{3.4} & - \\

    & UCB
    & \blue{\underline{0.3}} & \blue{0.9} & \blue{1.7} & \blue{0.9} & \blue{2.9} & - \\

    & InfoBatch 
    & \red{\textbf{0.0}} & \blue{\underline{0.5}} & \blue{\textbf{0.9}} & \red{\underline{0.0}} & \blue{\textbf{0.1}} & \blue{\textbf{1.7}} \\

    & Ours 
    & \red{\textbf{0.0}} & \blue{\textbf{0.3}} & \blue{1.6} & \red{\textbf{0.2}} & \blue{\underline{1.4}} & \blue{\underline{1.8}} \\
    
    % \midrule
    % \multicolumn{2}{c|}{Full Dataset} & \multicolumn{3}{c|}{-} & \multicolumn{3}{c}{-} \\
    \bottomrule
    \end{tabular}
    \end{adjustbox}
    \label{tab:pruning_comparison}
    \vspace{-0.095cm}
\end{table}

According to Table~\ref{tab:pruning_comparison}, our method outperforms all static pruning methods across both datasets and all pruning ratios, consistently achieving lower accuracy degradation. \changed{Compared} to dynamic methods, our approach delivers competitive performance, rivaling the state-of-the-art InfoBatch method~\citep{Qin:2024}. For example, on CIFAR-10 with a 50\% pruning ratio, our method results in an accuracy drop of only 0.3. Additionally, on CIFAR-100 with a 30\% pruning ratio, our method leads to a surprising increase in accuracy by 0.2, showcasing its robustness. 

Overall, the previous results emphasize the importance of timing in data pruning, demonstrating that applying these mechanisms after critical periods allows the model to retain more data during the most effective learning phases; thereby optimizing performance while reducing trade-offs.
% We apply our systematic approach using other architectures and datasets to compare with state of the art data pruning methods and validate our findings. This comparison is another reason we refrain from using the very best CIFAR10 models, like ResNet56, on the initial experiment. Table~\ref{tab:pruning_comparison} shows the results of this comparison.
%
% \changed{We apply our systematic approach using ResNet18 to compare the performance of pruning applied after the critical periods with other pruning methods that apply throughout the entire training process on CIFAR-10 and CIFAR-100. Our method initially trains on the full dataset for the first few epochs. From the critical periods onward, we apply the specified pruning ratio and continue training until completing 200 epochs. The baseline for calculating the accuracy delta is the model trained for the full 200 epochs on the entire dataset. The baseline and data for the other strategies are taken from  }

\noindent
\textbf{The Role of Annealing.}
\citet{Qin:2024} highlight that exposing the model to the full dataset during the final epochs, after applying data pruning, improves accuracy. This exposure allows the model to learn from data points that face exclusion by pruning. To achieve this, they introduce the Annealing parameter \( \delta \in (0, 1) \). This parameter controls when the model reverts to the full dataset, that is, we apply pruning only during the first \( \delta \cdot N \) epochs, where \( N \) is the total number of epochs in training. After that, we train the model on the complete dataset for the remaining epochs. Therefore, higher values of \( \delta \) lead to a greater reduction in training costs.

\begin{figure*}[h!]
	\centering
	\includegraphics[width=0.32\textwidth]{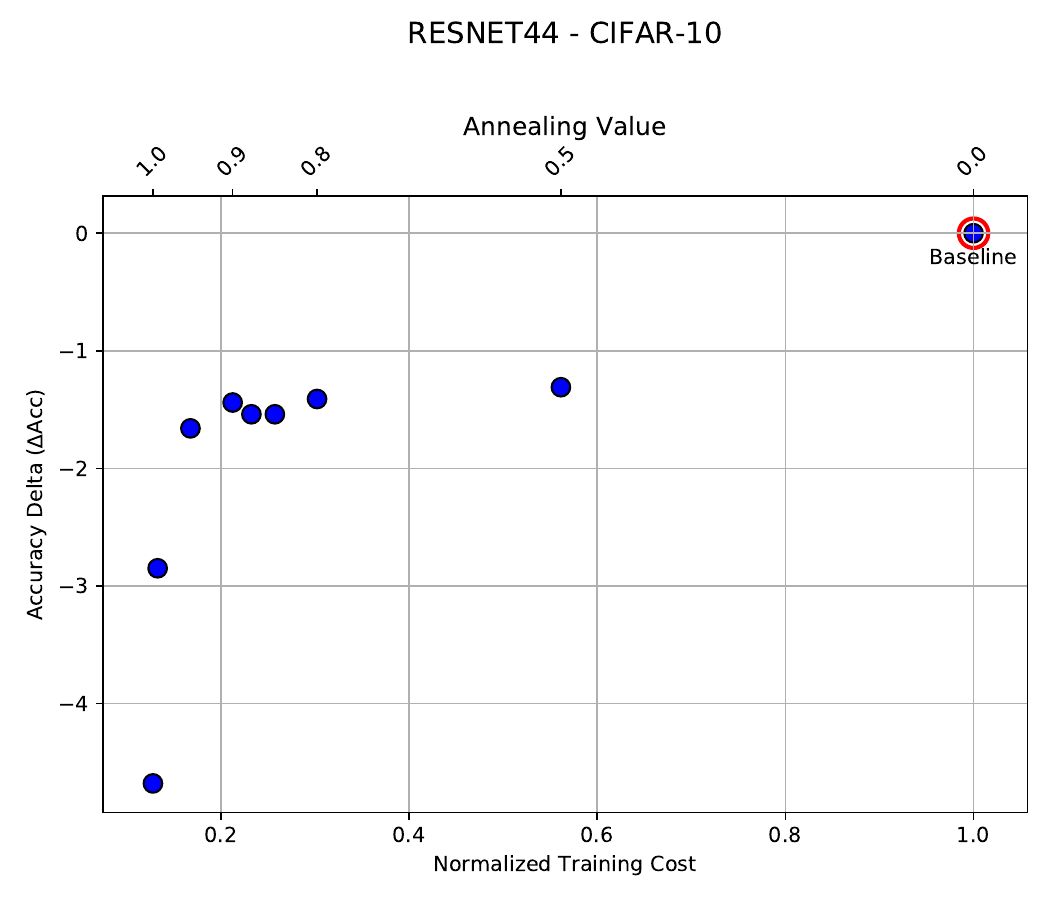}
	\includegraphics[width=0.32\textwidth]{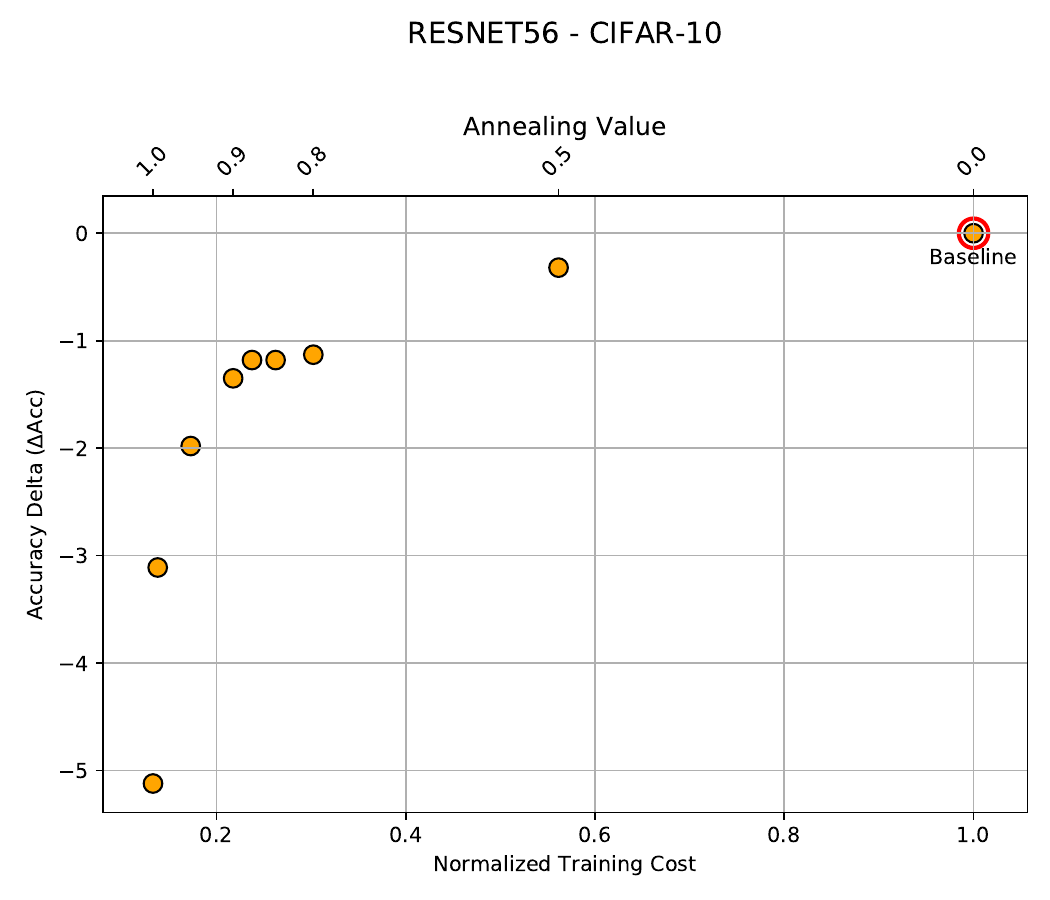} \hfill
	\includegraphics[width=0.32\textwidth]{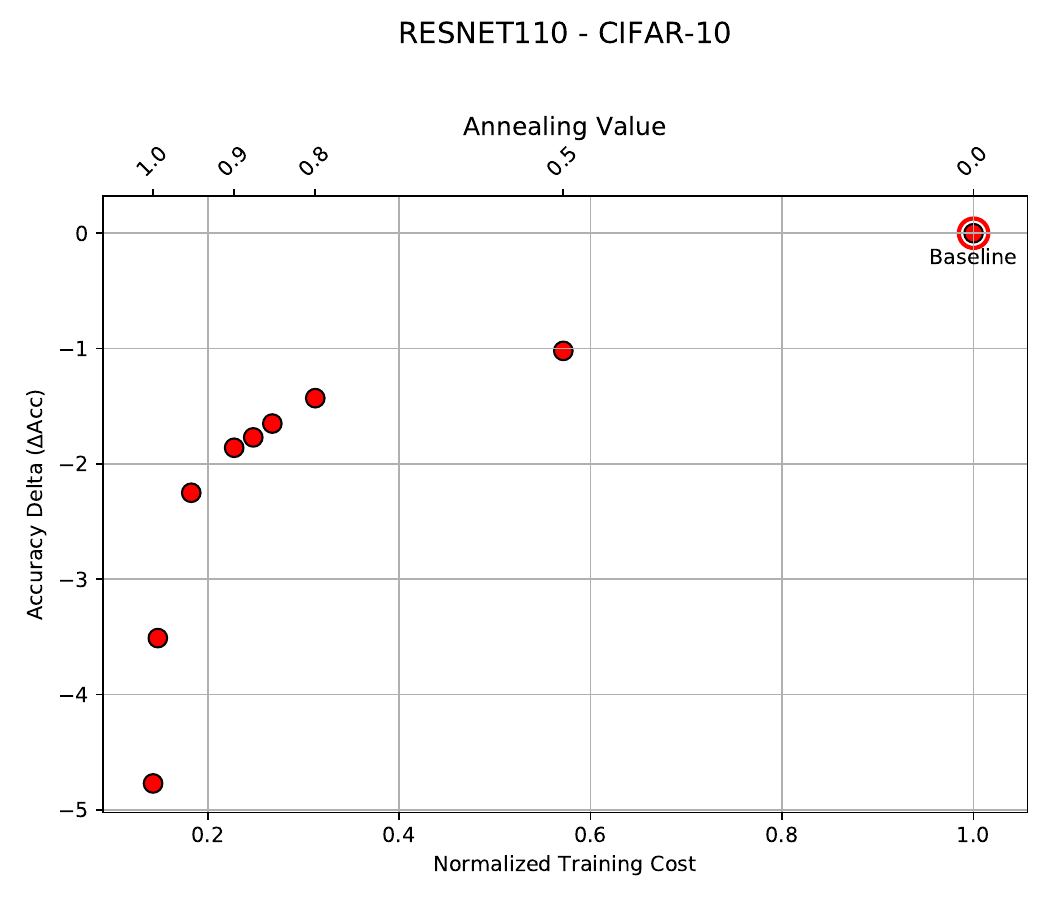} \\
	\includegraphics[width=0.32\textwidth]{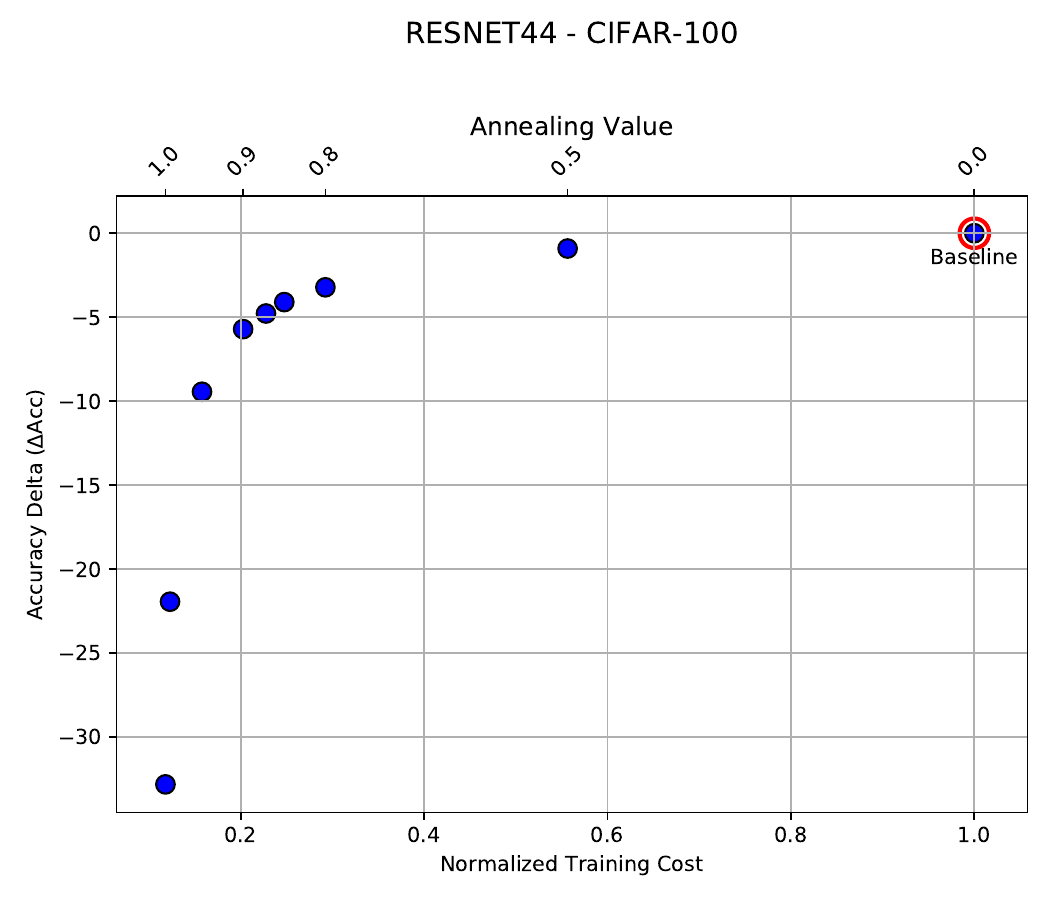}
	\includegraphics[width=0.32\textwidth]{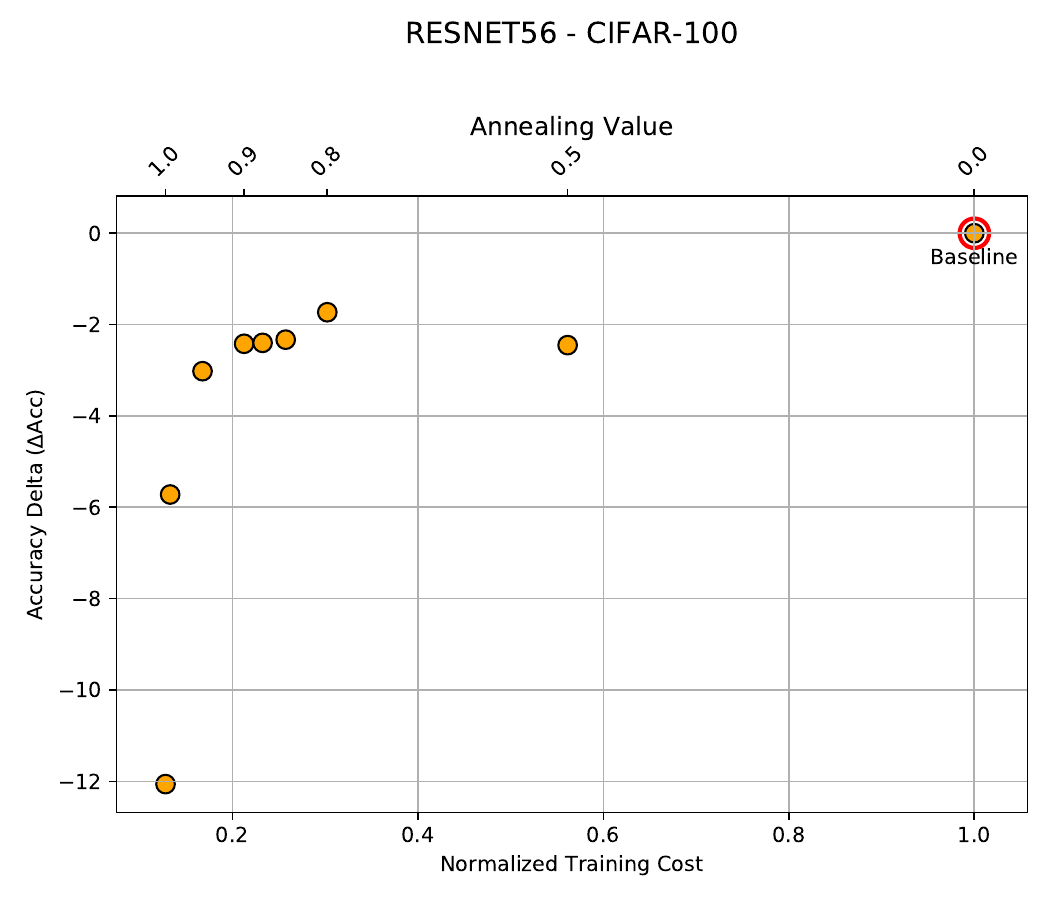} \hfill
	\includegraphics[width=0.32\textwidth]{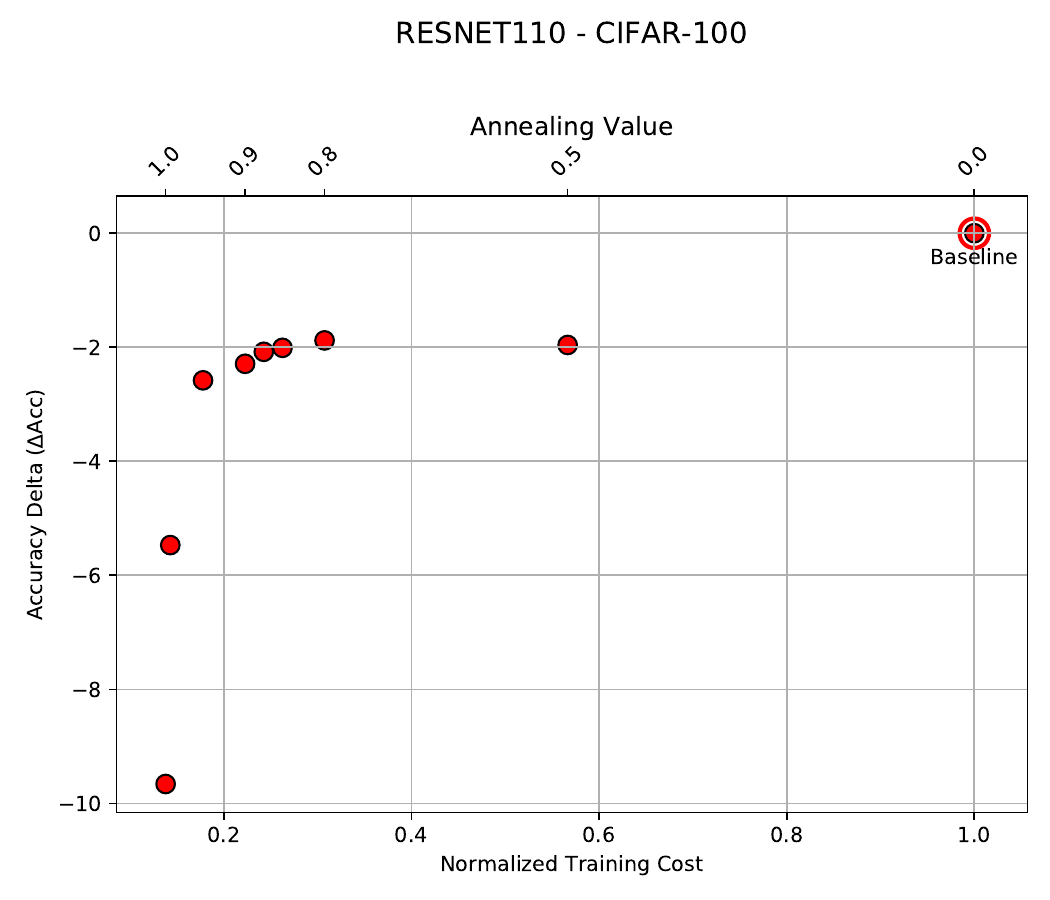}
	\caption{Impact of annealing on accuracy delta ($\Delta$Acc) in pp and training cost for different ResNet models and datasets (CIFAR-10/CIFAR-100). The training setup includes
	data augmentation with $k=3$ and data pruning with $k=0.01$.}
	\label{fig:graphs_annealing}
	\vspace{-0.095cm}
\end{figure*}

In this experiment, we combine the analysis of~\citet{Qin:2024}  with our contributions regarding critical periods to investigate the role of annealing in balancing training cost and model performance. To integrate both intensive training recipes and data pruning while accounting for critical periods, we first apply data augmentation (with \( k = 3 \)) until the critical period ends at epoch \( i^* \) found by our method. Following this, we introduce data pruning with a pruning ratio of 1\% (\( k = 0.01 \)) during the \( \delta \cdot (N - i^*) \) subsequent epochs. After this pruning phase, we resume the use of data augmentation until the conclusion of training.
%
% \changed{Following Qin et al.~\citep{Qin:2024},} the experiments also incorporate a parameter called Annealing. \changed{They highlight that exposing the model to the entire dataset during the final epochs, after applying data pruning, enhances accuracy. This exposure ensures the model captures knowledge from data points that may have been excluded by the pruning criteria. To control this effect, the Annealing parameter determines the number of final epochs during which the training data reverts to the initial proportion \( k \)  of the original dataset, used before the critical period.} It  directly influences the trade-off between computational efficiency and model performance, as captured by the Normalized Training Cost and Accuracy Delta metrics.
% For example, an annealing value of \( 1 \) signifies that, following the critical period, the dataset maintains the final \( k \) configuration throughout the remaining epochs. Conversely, an annealing value of \( 0.9 \) indicates that the first 90\% of the epochs after the critical period uses the final \( k \), while the last 10\% uses the initial \( k \). This dynamic adjustment allows investigating whether partial returns to the full dataset can recover lost accuracy without significantly increasing training costs.

Figure~\ref{fig:graphs_annealing} illustrates the impact of varying the annealing parameter, using $\delta$ values of  \( 1.0 \), \( 0.99 \), \( 0.95 \), \( 0.9 \), \( 0.875 \), \( 0.85 \), \( 0.8 \), and \( 0.5 \),  across different architectures for CIFAR-10/100. The results indicate that reverting to the full dataset for even a few epochs significantly improves accuracy while maintaining low normalized training costs. For example, with a lower annealing value such as \( 0.5 \), we reduce training costs by nearly 45\%, yet accuracy on the validation set remains close to the baseline. Notably, when \( N = 200 \) epochs, an annealing value of \( \delta = 0.99 \) means that only the final epoch employs data augmentation. Compared to the \( \delta = 1.0 \) (i.e., no annealing) configuration, this subtle change results in an accuracy improvement of up to 2 pp for CIFAR-10 and up to 11 pp for CIFAR-100, while incurring a negligible increase in training cost (less than 1 pp). These findings underscore the role of annealing in reintroducing the full dataset at the end of training to recover lost knowledge with minimal \changed{extra} cost.

By selecting an appropriate annealing value, one can balance normalized training cost and accuracy based on specific requirements, concentrating data points and computational effort during the critical period and at the final refinement stage, enabling the model to better capture and refine knowledge during training.

\noindent
\textbf{Effectiveness on Large Datasets and Optimizers.} 
To validate the effectiveness of our method, we analyze its performance across different datasets and optimizers. Our objective is to demonstrate that the method is agnostic to both dataset and optimizer choices, further reinforcing its general applicability. To this end, we consider the following datasets: CIFAR-10/100~\citep{cifar10,cifar100}, EuroSAT~\citep{helber2018eurosat}, TinyImageNet~\citep{Le2015tinyimagenet} and ImageNet30~\citep{hendrycks2019imagenet30}. On these datasets, the popular architecture is ResNet50; therefore, we employ it in these experiments.

Table~\ref{tab:dataset_variation} introduces the accuracy delta and time saved when applying our method on large datasets. The results confirm that our approach maintains competitive performance while significantly reducing computational costs, independent of the dataset. More concretely, we save up to \changed{65.79\%} of training time while increasing accuracy on \changed{EuroSAT by 1.02 pp.}

\begin{table}[h!]
	\centering
	\caption{Accuracy delta ($\Delta$Acc) \changed{in percentage points} and time saved of our method across different datasets using ResNet50 \changed{with augmentation factor $k=3$}. \changed{$\uparrow$ and $\downarrow$ mean accuracy improvement and decrease, respectively.}}
	\footnotesize
	\begin{tabular}{lccccc}
		\toprule
		& \textbf{CIFAR-10}             & \textbf{CIFAR-100} & \textbf{EuroSAT} & \textbf{TinyImageNet} & \textbf{ImageNet30} \\
		\midrule
		$\Delta$Acc &  $\downarrow$0.32 & $\downarrow$2.36   & $\uparrow$1.02   & $\downarrow$0.20      & $\uparrow$0.34 \\
		\midrule
		Saved (\%)  & 49.35        & 49.47       & 65.79      & 60.31       & 61.70 \\
		\bottomrule
	\end{tabular}
	\label{tab:dataset_variation}
	\vspace{-0.095cm}
\end{table}

Similarly, Table~\ref{tab:optimizer_variation} examines the impact of different optimizers on the effectiveness of our method in identifying critical periods. This experiment is important to show that our method is neither confined nor biased to the iterative learning process of SGD.
%
%\changed{The consistency across various optimizers demonstrate that our technique is optimizer-agnostic, ensuring its applicability in diverse training settings.}
Table~\ref{tab:optimizer_variation} confirms the consistency of our method across different optimizers, demonstrating its optimizer-agnostic nature and ensuring applicability in diverse training settings. \changed{The results indicate that changing the optimizer has minimal impact on our method. The accuracy deltas remain within a narrow range, and the time saved is consistently around 33\% across all optimizers, reinforcing the robustness of our approach.}

\begin{table}[h!]
	\centering
	\caption{Accuracy delta ($\Delta$Acc) \changed{in percentage points} and time saved of our method across different optimizers using ResNet50 \changed{on CIFAR-10 with augmentation factor $k=2$.} \changed{$\uparrow$ and $\downarrow$ mean accuracy improvement and decrease, respectively.}}
	\footnotesize
	\begin{tabular}{lcccc}
		\toprule
		& \textbf{SGD} & \textbf{AdamW} & \textbf{RMSprop} & \textbf{AdaGrad} \\
		\midrule
		$\Delta$Acc & $\downarrow$1.04 & $\downarrow$0.18 & $\downarrow$0.59 & $\downarrow$0.32 \\
		\midrule
		Saved (\%)  & 33.75       & 33.68       & 32.99       &  33.57 \\
		\bottomrule
	\end{tabular}
	\label{tab:optimizer_variation}
	\vspace{-0.095cm}
\end{table}

These results, combined with our previous experiments on different architectures, reinforce that our method is not only model-agnostic, as previously demonstrated, but also dataset-agnostic and optimizer-agnostic, further supporting its wide-ranging applicability.

\noindent
\textbf{Computational Benefits and GreenAI.} Existing works show that modern models emit high levels of carbon dioxide (CO$_2$) due to their substantial processing capacity and energy requirements during training and implementation~\citep{lacoste2019quantifying,faiz2024llmcarbon,morrison2025holistically}. However, our approach drastically lowers the carbon footprint through a direct increase in computational efficiency. Specifically, applying our method to ResNet56 results in a 58.89\% reduction in CO$_2$ emissions. Identifying critical periods also reduces financial costs, achieving a 58.33\% reduction \changed{for this model}. For other architectures, we achieve results nearing a 60\% reduction in both CO$_2$ emissions and financial costs. We estimate these values using the Machine Learning Impact Calculator~\citep{lacoste2019quantifying}. 

To summarize, our efforts yield significant advancements in GreenAI by effectively lowering carbon footprint and enhancing the financial accessibility of deep learning models.

\section{Conclusions}
Existing works suggest that early epochs, known as critical periods, play a decisive role in the success of many training recipes. In this work, we propose a systematic method to identify critical learning periods in neural network training.
\changed{Our method leverages a simple yet effective prediction estimator}, named Layer Rotation, to analyze the generalization behavior during training and then identify when critical periods emerge. 

Extensive experiments confirm a consistent pattern in our \review{idea}: larger layer rotations -- i.e., as cosine distance between the final and initial weights increases -- reliably predict enhanced generalization performance and hence indicate the emergence of critical periods.
Importantly, our method fills the gap in existing studies on critical learning periods that fail to offer ways to identify them. 
%From a practical perspective, our approach demonstrates significant efficiency improvements by \changed{X} regularization and data pruning strategies based on the critical period. 
From a practical perspective, our approach significantly improves training time by \changed{restricting resource-intensive training recipes (such as data augmentation) to the critical learning periods.}
As a concrete example, this results in up to $2.5\times$ \changed{reduction in training cost} with minimal accuracy trade-offs across diverse benchmarks. Our findings underscore the untapped potential of early-phase analysis for refining training recipes, offering practical insights for sustainable machine learning and resource allocation. We hope this work inspires further exploration into adaptive training methods and efficient deep learning practices.

\section*{Acknowledgments}
%FAPESP
% This study was financed, in part, by the São Paulo Research Foundation (FAPESP), Brasil. Process Number \#2023/11163-0. 
%CAPES
This study was financed in part by the Coordenação de Aperfeiçoamento de Pessoal de Nível Superior – Brasil (CAPES) – Finance Code 001. 
%CNPQ
% The authors would like to thank grant \#402734/2023-8, National Council for Scientific and Technological Development (CNPq). 
%Novos docentes
Artur Jordao Lima Correia would like to thank Edital Programa de Apoio a Novos Docentes 2023. Processo USP nº: 22.1.09345.01.2. 
Anna H. Reali Costa would like to thank grant \#312360/2023-1 CNPq.

\bibliographystyle{unsrtnat}
\bibliography{references}  %%% Uncomment this line and comment out the ``thebibliography'' section below to use the external .bib file (using bibtex) .




% 	\bibitem{keshet2016prediction}
% 	Keshet, Renato, Alina Maor, and George Kour.
% 	\newblock Prediction-Based, Prioritized Market-Share Insight Extraction.
% 	\newblock In {\em Advanced Data Mining and Applications (ADMA), 2016 12th International 
%                       Conference of}, pages 81--94,2016.

% \end{thebibliography}

\end{document}